\documentclass[letterpaper, 10 pt, conference]{ieeeconf}  

\IEEEoverridecommandlockouts                              

\usepackage{graphicx} 
\usepackage{float} 
\usepackage{subfigure} 
\usepackage{amsfonts,amssymb}
\usepackage{booktabs}
\usepackage{threeparttable}
\usepackage{multirow}

\title{\LARGE \bf
Point Cloud-Based Control Barrier Functions for Model Predictive Control in Safety-Critical Navigation of Autonomous Mobile Robots}

\author{Faduo Liang$^{1}$, Yunfeng Yang$^{2}$, and Shi-Lu Dai$^{1,*}$
\thanks{*This work was supported in part by the National Natural Science Foundation of China uunder Grant 42227901 and Grant 62473162, in part by the Guangdong Basic and Applied Basic Research Foundation under Grant 2024B1515120013, and in part by the Guangdong Emergency Management Science and Technology Program under Grant 2025YJKY002. \emph{(Corresponding author: Shi-Lu Dai.)}}
\thanks{$^{1}$Faduo Liang and Shi-Lu Dai are with the School of Automation Science and Engineering, South China University of Technology, Guangzhou 510641, China, and also with the Southern Marine Science and Engineering Guangdong Laboratory (Zhuhai), Zhuhai 519000, China
        {\tt\small aulfd@mail.scut.edu.cn; audaisl@scut.edu.cn}}%
\thanks{$^{2}$ Yunfeng Yang is with the Guangdong Academy of Safety Production and Emergency Management Science and Technology, Guangzhou, 510060, China
        {\tt\small yunfengyang1990@163.com}}%
\thanks{Code: \href{https://github.com/Faduo-Liang/PCCBF-MPC}{https://github.com/Faduo-Liang/PCCBF-MPC}.}
\thanks{The demonstrated video of the experiment may be found in the attached documents.}
}

\begin{document}

\maketitle
\thispagestyle{empty}
\pagestyle{empty}

\begin{abstract}
In this work, we propose a novel motion planning algorithm to facilitate safety-critical navigation for autonomous mobile robots. The proposed algorithm integrates a real-time dynamic obstacle tracking and mapping system that categorizes point clouds into dynamic and static components. For dynamic point clouds, the Kalman filter is employed to estimate and predict their motion states. Based on these predictions, we extrapolate the future states of dynamic point clouds, which are subsequently merged with static point clouds to construct the forward-time-domain (FTD) map. By combining control barrier functions (CBFs) with nonlinear model predictive control, the proposed algorithm enables the robot to effectively avoid both static and dynamic obstacles. The CBF constraints are formulated based on risk points identified through collision detection between the predicted future states and the FTD map. Experimental results from both simulated and real-world scenarios demonstrate the efficacy of the proposed algorithm in complex environments. In simulation experiments, the proposed algorithm is compared with two baseline approaches, showing superior performance in terms of safety and robustness in obstacle avoidance. The source code is released for the reference of the robotics community.
\end{abstract}

\section{INTRODUCTION}
Research on safety-critical optimal planning and control for autonomous mobile robots has been actively pursued over the past decades. Significant progress has been made in collision avoidance methods for both static and dynamic environments, with related algorithms becoming increasingly mature. However, reliable and efficient collision avoidance in environments containing both static (e.g., tables, shelves) and dynamic (e.g., pedestrians) obstacles remains a significant challenge.

A primary challenge in mobile robot navigation lies in accurately characterizing obstacles. Existing works \cite{MPCC, MPC-DCBF, CLF-CBF, AF} typically model obstacles as ellipsoids through clustering and generate optimal trajectories under obstacle avoidance constraints. However, such approaches can lead to a certain degree of conservativeness \cite{Sailing-cbf}. Firstly, density-based clustering methods like DBSCAN \cite{DBSCAN} are sensitive to variations in local data density and require careful tuning of parameters (e.g., eps and minPts), limiting their adaptability to high-dimensional spaces \cite{AF}. Secondly, representing obstacles as ellipsoids reduces the available state space for robots \cite{Onboard-Mapping}. This approach overestimates obstacle volumes by including irrelevant regions, such as hollow spaces, particularly for irregularly shaped obstacles, which restricts robot movement. Such over-approximation significantly impacts the feasibility and optimality of motion planning in complex environments. Recent studies \cite{Sailing-cbf, PCCBF, ReactiveCA} have proposed several important methods to characterize obstacles using point cloud data, but these methods still struggle to handle complex and dynamic environments effectively.

\begin{figure}
\centering
\includegraphics[width=0.48\textwidth]{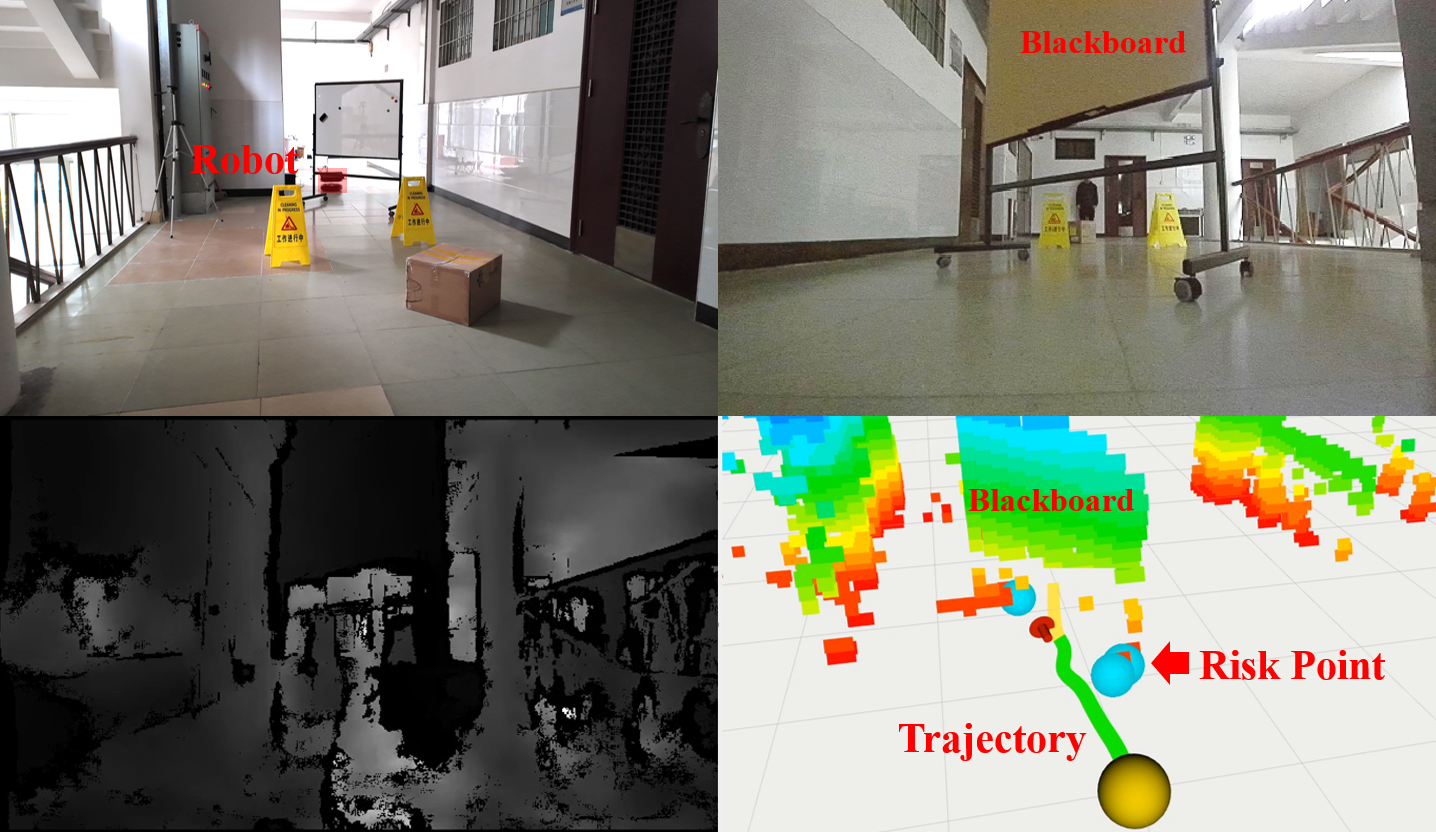}
\caption{Experimental results of navigation in an indoor environment with onboard sensing. Top left: an overview of the experimental setup and trajectory results. Top right: onboard color image. Bottom left: depth image. Bottom right: the optimized trajectory (green curve) through the hollow area inside the blackboard.}
\label{Fig.1}
\end{figure}

\begin{figure*}[htbp]
\centering
\includegraphics[width=1\textwidth]{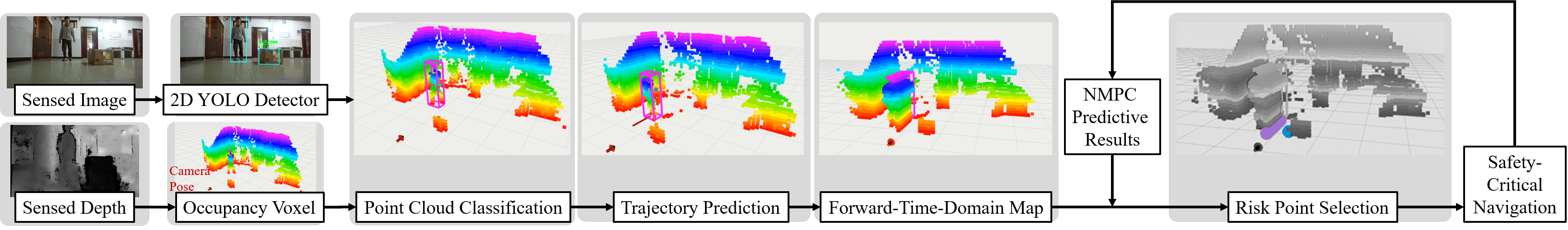}
\caption{The proposed collision avoidance framework. The perception pipeline processes sensory data into occupancy voxel map. This occupancy voxel map is subsequently utilized to compute a forward-time-domain (FTD) map, which identifies static and dynamic risk points based on the nonlinear model predictive control (NMPC) prediction results. The points with the highest risk of collision (static risk points in blue and dynamic risk points in purple) are identified through collision detection. These high-risk points dynamically define safety constraints within the NMPC framework via control barrier functions, enabling safe collision-free navigation.}
\label{Fig.2}
\end{figure*}

Jointly solving motion planning and control under perception constraints also presents substantial difficulties. Nonlinear model predictive control (NMPC) addresses these challenges by integrating planning and control into a unified constraint optimization framework \cite{MPC-DCBF}. However, despite its theoretical appeal, most approaches primarily use NMPC as an optimal controller, underutilizing its predictive capabilities and limiting their effectiveness in real-world applications. Recently, control barrier functions (CBFs) have emerged as a widely adopted method for enforcing safety constraints in obstacle avoidance. In \cite{PCCBF}, surface geometry was employed to locally define and synthesize a quadratic CBF over point cloud data. A MPC-DCBF framework was presented in \cite{MPC-DCBF} to address dynamic obstacle avoidance. However, these existing CBF-based methods still face challenges in simultaneously handling both dynamic and static obstacles.

To address these challenges, we propose a novel framework that integrates NMPC with point cloud-based CBFs to simultaneously handle both static and dynamic obstacles under real-time perception constraints. The contributions of our work are summarized as follows:

\begin{enumerate}
   \item[1)] We develop a point cloud-based CBF-NMPC algorithm that integrates planning and control, enabling robots to generate safe and collision-free trajectories in unknown and dynamic environments.
  \item[2)] We propose a heuristic method for identifying high-risk collision points using the forward-time-domain (FTD) map and the predictive capabilities of NMPC. This method effectively characterizes potential collisions and formulates CBFs for enhanced safety.
  \item [3)]We conduct experiments in both Gazebo simulations and real-world scenarios. The results demonstrate the effectiveness and robustness of our safety-critical obstacle avoidance algorithm in unstructured environments with static and dynamic obstacles.
\end{enumerate}

\section{RELATED WORK}
\subsection{Dynamic Perception}
Obstacle detection and tracking methods can be categorized into two groups based on the input data representations.

\emph{Image-Based Algorithms:}
In \cite{detect1, detect2, detect3}, depth images are utilized to generate U-depth and V-depth maps, enabling the classification of obstacles and the estimation of their states. However, these methods often represent static obstacles as ellipsoids, which can be overly conservative for complex-shaped obstacles \cite{detect2}. In contrast to prior depth image-based approaches, a learning-based detector \cite{Leveraging} is introduced to improve detection accuracy.

\emph{Point Cloud-Based Algorithms:}
Detection algorithms based on iterative closest point \cite{ICP} match current point cloud segments to previous ones to estimate motion. Although effective for rigid objects like cars, this approach performs poorly for deformable objects such as humans, making it unsuitable for systems that prioritize human detection and tracking. A dynamic environment perception method was proposed in \cite{AF}, where dynamic obstacles were modeled as ellipsoids using RGB-D camera data. In \cite{Leveraging}, a point cloud clustering method was combined with a YOLO detector for human detection.

Both image-based and point cloud-based algorithms can suffer from misdetection due to noise and environmental complexity. To address this, recent research \cite{Onboard-Detect} has focused on ensemble methods to mitigate their individual shortcomings. Our approach builds upon and extends the dynamic obstacle detection methods \cite{Onboard-Detect}. Specifically, we replace the human detector with our proposed YOLO-Fusion detector to reduce 3D bounding box estimation errors.

\subsection{Control Barrier Function-Based Collision Avoidance}
In recent years, control barrier functions (CBFs) \cite{MPC-DCBF, CLF-CBF, Sailing-cbf, PCCBF, MPC-CBF, CBF, Lyapunov-Based-cbf, Learning-CBF} have gained significant attention for ensuring set invariance by considering system dynamics and guaranteeing forward invariance of the safe set. A control methodology was presented in \cite{CBF} to unify CBFs and control Lyapunov functions (CLF) through quadratic programs (QP). A novel CLF-CBF-QP obstacle avoidance system \cite{CLF-CBF} was designed for bipedal robots navigating among multiple non-overlapping obstacles. However, these methods rely on current state information without prediction, resulting in a greedy and short-sighted control policy.

In contrast, CBF-NMPC incorporates future state information to improve performance. A safety-critical model predictive control (MPC) strategy was presented in \cite{MPC-CBF} to employ discrete-time CBFs, which guarantees system safety and yields a less greedy control policy. A  MPC-DCBF framework \cite{MPC-DCBF} was developed for
dynamic obstacle avoidance. However, these methods formulate CBFs by encapsulating obstacles with ellipsoids and representing the robot as a sphere, potentially leading to over-conservatism.

The synthesis of CBFs is one of the main challenges in utilizing CBF-based methods for navigation \cite{Learning-bCBF}.  An online learning-based CBF controller were designed in \cite{Learning-CBF} for static environments by stereo depth data. \cite{PCCBF} presents a computationally low-cost framework for synthesizing barrier functions over point cloud data for safe vision-based control. However, these methods may suffer from collision issues in complex environments. To address this, we propose a CBF synthesis method that generates CBFs using a series of historical, current, and future risk points in the FTD map, enabling far-sighted and robust collision avoidance.

\subsection{Model Predictive Control-Based Trajectory Planning}
A integrated planning and control framework was presented in \cite{IPC}, which relies on generating safe flight corridors to represent free space in the environment and employs multiple hyperplanes of polyhedrons as linear inequality constraints in the MPC problem to ensure flight safety. The integrated planning and control framework avoids dynamic objects in a reactive manner without tracking or predicting object movement. However, such reactive obstacle avoidance is fundamentally limited by the maneuverability of the quadrotor. Planned obstacle avoidance, which involves detecting, tracking, and predicting the movement of dynamic objects, could overcome this limitation and enable the avoidance of higher-speed moving objects. In \cite{on-demand}  a trajectory generation method introduced based on MPC prediction, and on-demand collision avoidance was incorporated within a distributed MPC framework. This approach detects and resolves only the first collision within the finite prediction horizon, reducing computation time and increasing the success rate for transition tasks. Compared to \cite{MPC-DCBF, MPC-CBF}, which include all obstacles in the environment as constraints and utilize MPC prediction for dynamic obstacle avoidance, our approach extends on-demand collision avoidance to reduce computational load and enable real-time navigation in more general environments.

\section{DYNAMIC ENVIRONMENT PERCEPTION, PLANNING AND CONTROL}
The objective is to enable an autonomous robot equipped with an RGB-D camera to navigate safely from start to goal in unknown environments containing static and dynamic obstacles. The robot must perceive its surroundings in real time and make decisions to avoid collisions while reaching its goal, without prior knowledge of the environment, such as the shape or state of the obstacles.

The algorithm flow of the system is illustrated in Fig. \ref{Fig.2}. The sensed depth data from the RGB-D camera is converted into an occupancy voxel map, which is then classified into dynamic and static point clouds. Using a Kalman filter, the current motion state of dynamic point clouds is estimated and used to generate the corresponding trajectory in the forward time domain. The predicted trajectory is sent to build the FTD map to identify risk points combined with the predicted state results from NMPC. These risk points are integrated into the NMPC optimization via CBFs, allowing the controller to enforce safety constraints during navigation.

\subsection{Dynamic Obstacle Detection} \label{III.A}
The perception module processes the depth image to generate raw point cloud data, which is then converted into an occupancy voxel map by applying the voxel filter proposed in \cite{Leveraging} to remove the noise from the point cloud. We define the point cloud from occupancy voxel map as $P_{static}=\{ (x_i, y_i, z_i)\}_{i=0}^{M_s}$ in the spatial frame.

\begin{figure}[htbp]
\centering
\includegraphics[width=0.48\textwidth]{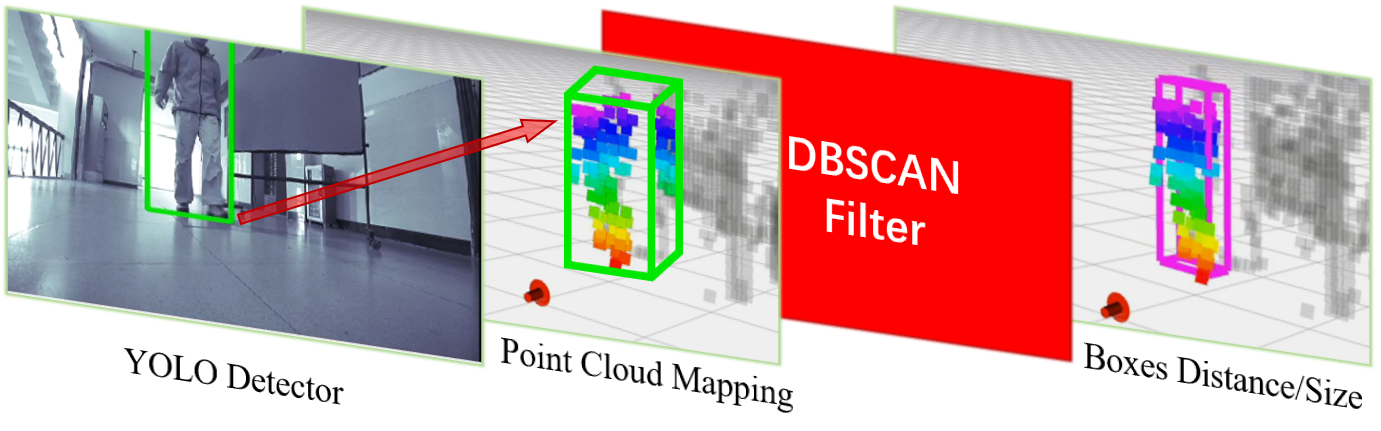}
\caption{Illustration of the YOLO-Fusion detector. The RGB image is first processed to obtain the 2D YOLO detection results, from which the corresponding bounding box on the depth image is derived. Using this 2D bounding box information, the proposed filtering method calculates the 3D bounding box for accurate object localization.}
\label{Fig.3}
\end{figure}

Pedestrians are common dynamic obstacles in indoor environments for robots. We introduce our YOLO-Fusion detector, which is based on 2D YOLOFastestDet and can run in real-time on an onboard CPU. The architecture of the YOLO-Fusion detector is illustrated in Fig. \ref{Fig.3}. Unlike traditional image-based YOLO detectors, the YOLO-Fusion detector combines image data and point cloud data to achieve precise and stable 3D detection results, including depth and thickness. The detector first identifies the 2D bounding box of each pedestrian in the RGB image and then extracts the corresponding point clouds from the occupancy voxel map. To remove unrelated point clouds, we utilize the DBSCAN algorithm as a point cloud filter. This algorithm clusters the corresponding point clouds and selects the largest cluster as the detection result of the YOLO-Fusion detector. Finally, the size and position of the 3D obstacle's bounding box can be easily obtained from the filtered point clouds. Unlike \cite{Leveraging}, our approach does not require clustering all point clouds in the environment, thereby reducing unnecessary computational overhead.

Since this learning-based detector may still lack generalization ability, we combine our YOLO-Fusion detector with the robust detection framework \cite{Onboard-Detect}, which integrates U-depth and DBSCAN detectors to identify universal dynamic obstacles.
\subsection{Local Map Construction} \label{III.B}
In this module, we aim to construct an FTD map that contains predictive information for the next $N$ time steps. As dynamic obstacles are identified, their corresponding point clouds are removed from the static occupancy voxel map $P_{static}=\{ (x_i, y_i, z_i)\}_{i=0}^{M_s}$ and added to the dynamic map $P_{dyn,0}=\{ (x_{i0}, y_{i0}, z_{i0})\}_{i=0}^{M_d}$. Using a constant velocity model, a Kalman filter (KF) is employed to estimate and predict the motion state of the dynamic point cloud $P_{dyn,k}$, resulting in $P_{dyn,k}=\{ (x_{ik}, y_{ik}, z_{ik})\}_{i=0}^{M_d}$, $k=0,1,...,N-1$. The FTD map is described by
$$
P_{FTD}=P_{static} \cup P_{dyn,k},~~ k=0,1,...,N-1, \eqno{(1)}
$$
which contains the point cloud at the current time step and the predicted point clouds for the next $N$ time steps.

\begin{figure}[htbp]
\centering
\includegraphics[width=0.48\textwidth]{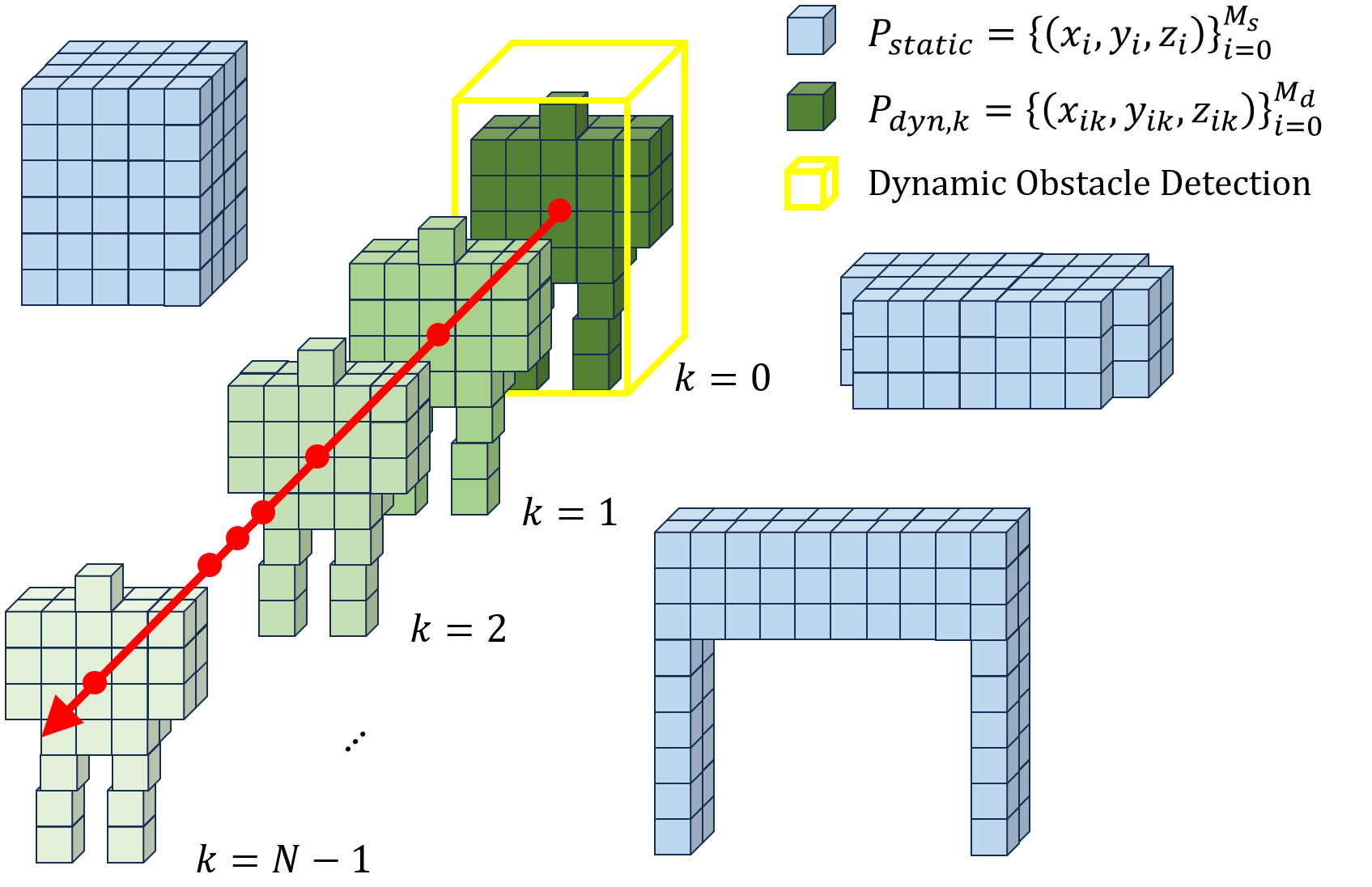}
\caption{FTD map structure.}
\label{Fig.4}
\end{figure}

To enhance query efficiency for risk point identification (Section \ref{III.E}) and enable efficient collision checking, the local map is organized into $N+1$ KD-Trees: one for the static map $P_{static}$ and $N$ for the dynamic map $P_{dyn,t}$ at each time step. The structure of the FTD map is shown in Fig. \ref{Fig.4}.

\subsection{CBF Generation} \label{III.C}
CBFs have been proven effective for obstacle avoidance in unmanned ground vehicles, e.g., \cite{MPC-DCBF, PCCBF, MPC-CBF}. For safety-critical control, we define a set $\mathcal{C}$ as the superlevel set of a continuously differentiable function $h : \mathcal{X} \subset \mathbb{R}^n \to \mathbb{R} $:
$$
\mathcal{C}=\{\mathrm{x}\in \mathcal{X}\subset \mathbb{R}^n:h(\mathrm{x})\geq 0\}, \eqno{(2)}
$$
where $\mathcal{C}$ is referred to as a \emph{safe set}. The function $h$ is a CBF \cite{CBF} if $\frac{\partial h}{\partial x} \neq 0$ for all $x \in \partial \mathcal{C}$ and there exists an extended class $\mathcal{K}_\infty$ function $\gamma$, $h$ satisfies
$$
\exists\mathrm{u}\ \textrm{s.t.}\ \dot{h}(\mathrm{x},\ \mathrm{u})\geq-\gamma(h(\mathrm{x})),~~ \gamma\in \mathcal{K}_{\infty}. \eqno{(3)}
$$
Condition (3) can be extended to the discrete-time domain as follows
$$
\Delta h(\mathrm{x}_{k}, \mathrm{u}_{k})\geq-\gamma h(\mathrm{x}_{k}),~~\ 0 < \gamma\leq 1, \eqno{(4)}
$$
where $\Delta h(\mathrm{x}_k, \mathrm{u}_k) := h(\mathrm{x}_{k+1}) - h(\mathrm{x}_k)$. By enforcing constraint (4), we derive $h(\mathrm{x}_{k+1}) \ge (1 - \gamma)h(\mathrm{x}_k)$, indicating that the lower bound of $h(\mathrm{x})$ decreases exponentially at a rate of $1-\gamma$ \cite{MPC-CBF}. According to the main CBF theorem, the safe set $\mathcal{C}$ is forward invariant and asymptotically stable \cite{CBF_theory}. Therefore, the robot position within $\mathcal{C}$ is also invariant and asymptotically stable, implying that the control system remains safe.

To ensure the system maintains a safe distance from potential collisions with the environment, we define the safe set $\mathcal{C}$ based on the system's knowledge of the environment, represented by a point cloud $P_{FTD} = \{(x_i, y_i, z_i)\}^{M_{FTD}}_{i=1}$ in the spatial frame. The safety condition is defined using the barrier function:
$$
h(\mathrm{x}_k)=\mathop{\min}\limits_{\mathrm{x}_p\in P_{FTD}}\ \{ (\mathrm{x}_k - \mathrm{x}_p)^T(\mathrm{x}_k - \mathrm{x}_p)-\delta^2 \} \eqno{(5)}
$$
where $\delta > 0$ is a buffer that defines the minimum safe distance to the environment. Since $h(x_k)$ considers the minimum distance from the system to the point cloud, $h(x_k) > 0$ implies that the system is not in collision with any point in the point cloud.

To employ the barrier functions to the NMPC framework, we rewrite it as:
$$
h(\mathrm{x}_k)= (\mathrm{x}_k - \mathrm{x}^*_p)^T(\mathrm{x}_k - \mathrm{x}^*_p)-\delta^2,~~ k=0,\ldots,N-1 \eqno{(6)}
$$
where $\mathrm{x}^*_p$ represents the risk points in the FTD map relative to the system's position $\mathrm{x}_k$ at horizon step $k$.

\subsection{MPC-Based Planning and Control} \label{III.D}
The objective of MPC is to guide the robot to the target location while keeping it away from risk points represented by CBFs (Section \ref{III.E}) and satisfying necessary physical constraints. At every discrete-time index $t$, the robot computes a new input sequence over the horizon through the following three steps:
\begin{enumerate}
  \item \emph{Collision Checking:} Detect potential future collisions using the latest predicted states computed at time $t-1$ of the robot and the FTD map.
    \item \emph{Optimization Problem Construction:} Formulate the optimization problem, including state and actuation constraints, and add collision constraints only if necessary.
  \item \emph{Control Execution:} Apply the first element of the obtained optimal input sequence to the robot.
\end{enumerate}

Predicting collisions is the core idea behind risk point identification in CBF generation (Section \ref{III.E}). This process repeats until the robot reaches the desired objective.

The robot's prediction model is described by the following discrete-time equation
$\mathbf{x}_{t+1}=f(\mathbf{x}_{t},\mathbf{u}_{t})$, where $\mathbf{x}_t \in \mathcal{X} \subset \mathbb{R}_n$ represents the states of the system at time step $t \in \mathbb{Z}^+$, and $\mathbf{u}_t \in \mathcal{U} \subset \mathbb{R}_m$ is the control inputs. Here, $f(\cdot)$ is locally Lipschitz as the differential-drive system model:
$$
\mathbf{x}_{t+1}=\mathbf{x}_t+
\left[
\begin{array}{ccc}
    \cos \theta_t & 0 \\
    \sin \theta_t & 0 \\
    0 & 1
\end{array}
\right]\mathbf{u}_t {\Delta t},
\eqno{(7)}
$$
with $\mathbf{x}_t=\left[ x_t,y_t,\theta_t \right]$, where $\theta_t$ represents the robot direction.

We sample \emph{N}-step horizon for the MPC and set the target location $\mathbf{x}_{tar}$ as the goal of robot. The MPC problem is formulated as:
$$
\mathop{\min}\limits_{\mathbf{u}_{t:t+N-1\vert t}} p(\mathbf{x}_{t+N\vert t}) + \sum\limits_{k=0}^{N-1}q(\mathbf{x}_{t+k\vert t},\mathbf{u}_{t+k\vert t}) \eqno{(8a)}
$$
$$
\textrm{s.t.} \ \mathbf{x}_{t+k+1\vert t}=f(\mathbf{x}_{t+k\vert t},\mathbf{u}_{t+k\vert t}), \eqno{(8b)}
$$
$$
\mathbf{x}_{t+k\vert t} \in \mathcal{X}, ~~\mathbf{u}_{t+k\vert t} \in \mathcal{U}, \eqno{(8c)}
$$
$$
\mathbf{x}_{t|t} = \mathbf{x}_t, \eqno{(8d)}
$$
$$
\mathbf{x}_{t+N|t} \in \mathcal{X}_f, \eqno{(8e)}
$$
$$
\Delta h(\mathrm{x}_{t+k\vert t},\mathrm{u}_{t+k\vert t})\geq-\gamma h(\mathrm{x}_{t+k\vert t}),~~ k=0, \ldots, N-1. \eqno{(8f)}
$$

\begin{figure}[htbp]
\centering
\includegraphics[width=0.48\textwidth]{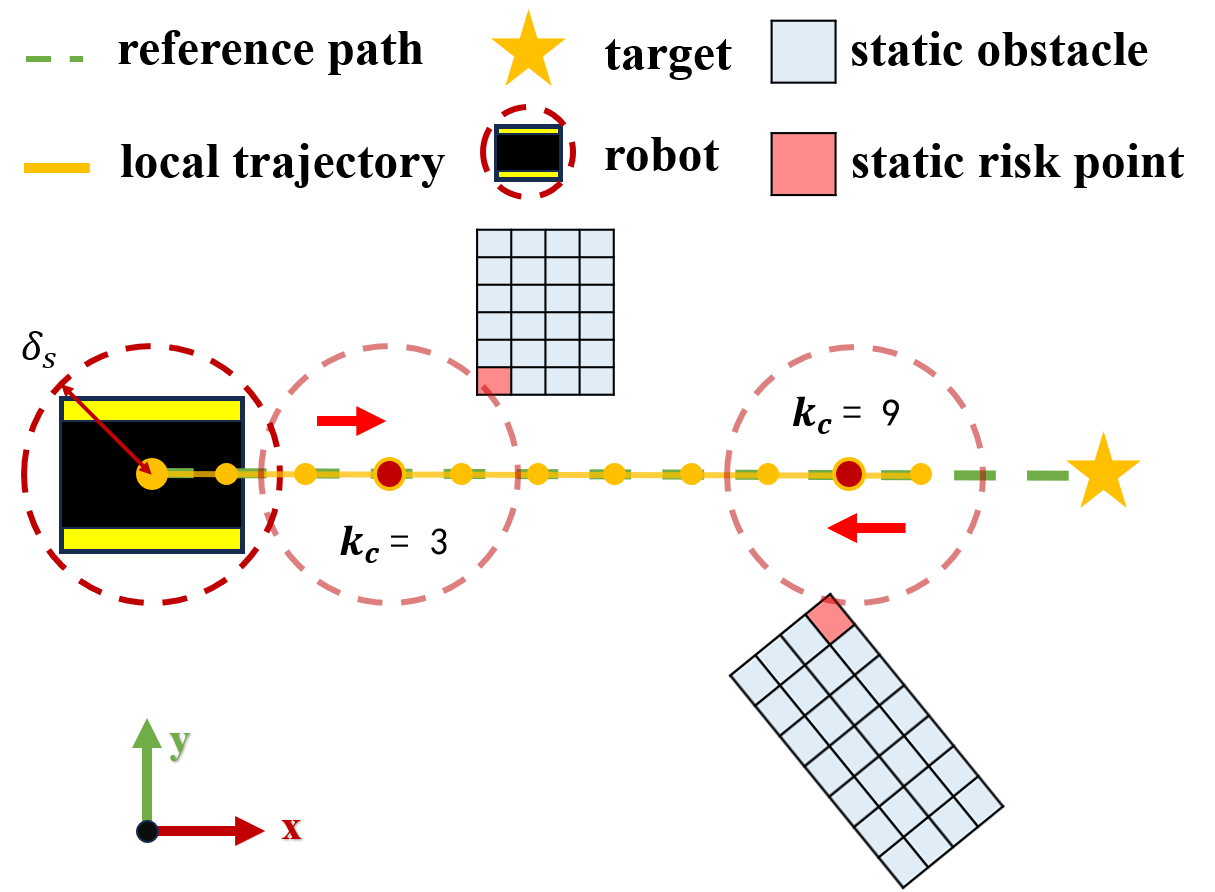}
    \caption{Static risk point identification methodology. Forward detection and inverse collision detection leverage the FTD map data and NMPC predictions to identify static risk points.}
\label{Fig.5}
\end{figure}

The cost function (8a) consists of stage cost $q(\mathbf{x}_{t+k|t},\mathbf{u}_{t+k|t})=\Vert \mathbf{x}_{tar}-\mathbf{x}_{t+k|t} \Vert^2_{Q} + \Vert \mathbf{u}_{t+k|t} \Vert^2_{R} + \Vert \mathbf{x}_{t+k|t-1}-\mathbf{x}_{t+k|t} \Vert^2_{W}$, which represents target tracking error, control effort, and prediction variation error, respectively, and terminal cost $p(\mathbf{x}_{t+N|t})=\Vert \mathbf{x}_{tar}-\mathbf{x}_{t+N|t} \Vert^2_{Q_N}$, which represents terminal target tracking error. Here, $Q$, $R$, $W$, and $Q_N$ are weight matrices, while $\mathcal{X}$, $\mathcal{U}$, and $\mathcal{X}_f$ represent the sets of admissible states, inputs, and terminal states, respectively. Note that incorporating the prediction variation error enables the robot to mitigate planning disturbances.

The MPC problem (8) includes four constraints. The first constraint is the model dynamics (8b), subject to the initial states (8d) estimated by odometry. The second constraint (8c) ensures the system states and inputs remain within feasible domains. The third constraint (8e) enforces the terminal set constraint. The fourth constraint incorporates the CBF constraints (8f), ensuring the robot remains in a safe state, avoiding collisions with dynamic and static risk points.

When the optimization problem (8) is successfully solved at time $t$, the optimal solution $\mathbf{u}_{t:t+N-1|t}^*=\{ \mathbf{u}_{t|t}^*, \ldots, \mathbf{u}_{t+N-1|t}^* \}$ is obtained. The first element $\mathbf{u}_{t|t}^*$ is applied as the control action. If the MPC fails at time $t$, the time interval $\Delta T$ between the last successful step and the failed time $t$ is calculated. The control action $\mathbf{u}_{t+\Delta T|t-\Delta T}^*$ from the previously successful solution $\mathbf{u}_{t:t+N-1|t-\Delta T}^*$ is used instead. This strategy ensures adaptability to MPC failures and guarantees a reasonable control action even when the MPC cannot be solved.

\begin{figure}[htbp]
\centering
\includegraphics[width=0.48\textwidth]{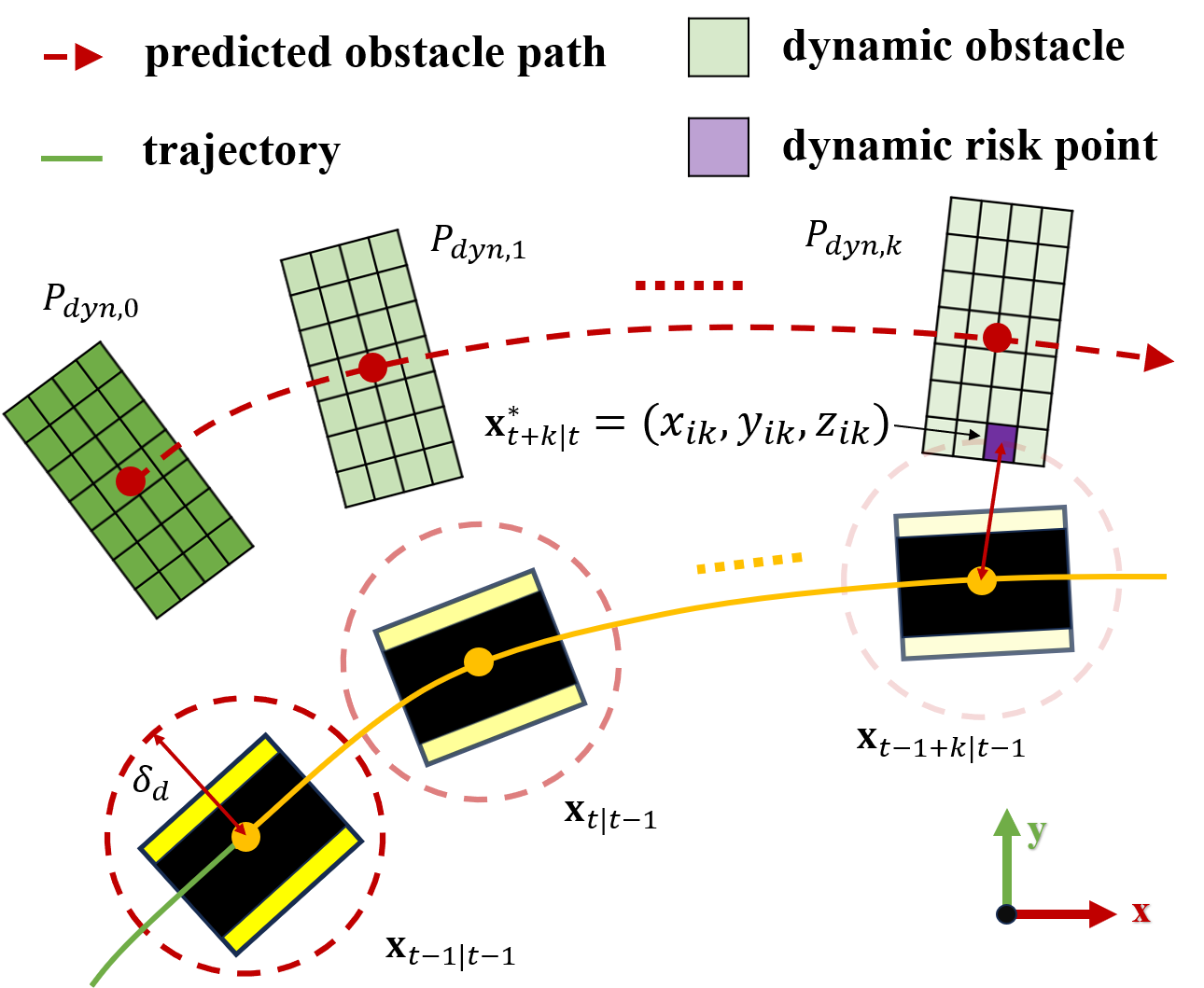}
\caption{Dynamic risk point identification methodology. At solving time $t$, the predicted path of the robot $\mathbf{x}_{t-1:t-1+N-1|t-1}$ is determined to identify dynamic risk point $\mathbf{x}_{t+k|t}^*$ with dynamic obstacle $P_{dyn,k}$ for $k=0,1,\dots,N-1$.}
\label{Fig.6}
\end{figure}

\begin{figure*}[htbp]
\centering
\includegraphics[width=1\textwidth]{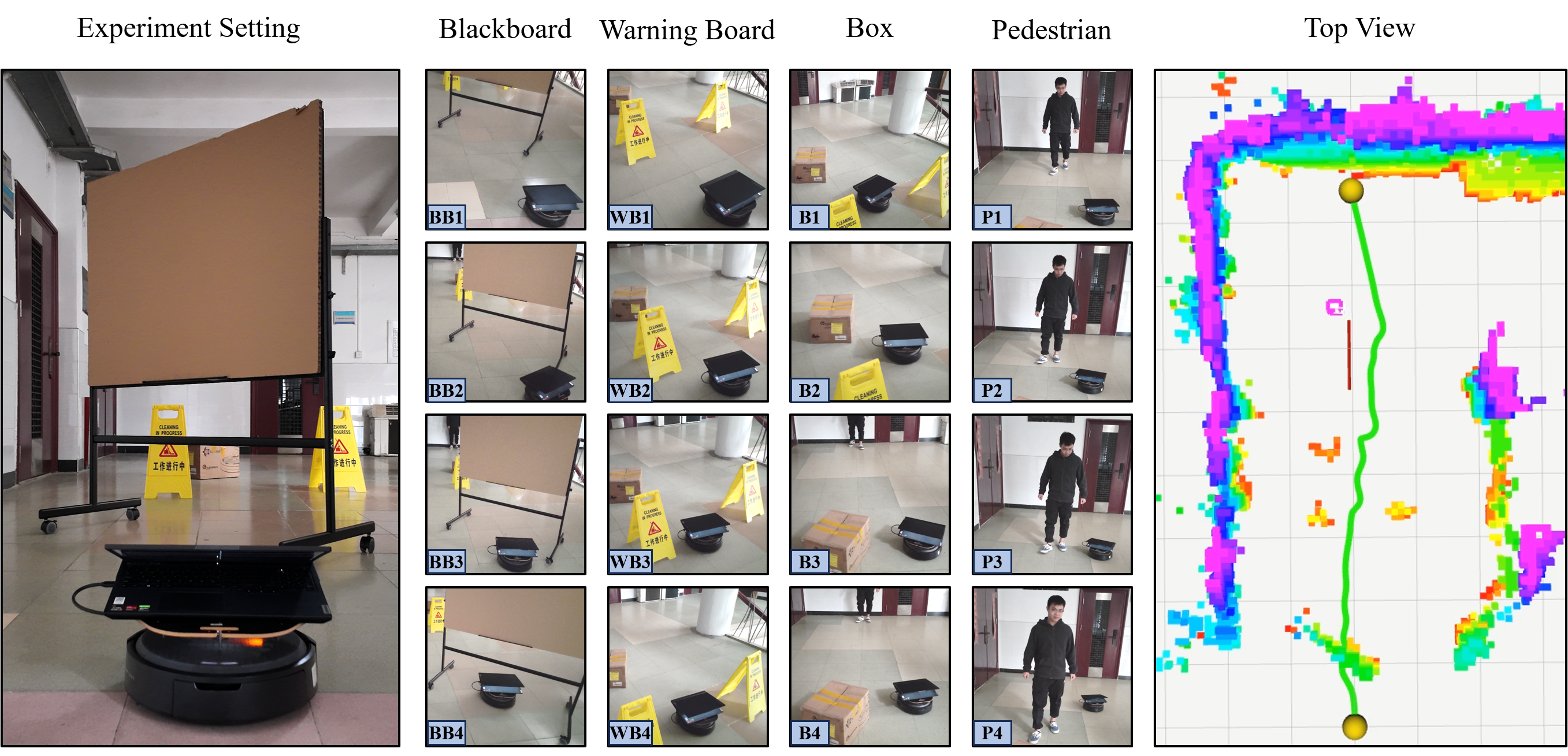}
\caption{Real-world experimental results. The illustrated settings encompass four principal obstacles. For each obstacle, four snapshots are provided to illustrate the motion generated by the proposed local planner. The right figure delineates the trajectory of our method, with the map color-coded according to height.}
\label{Fig.7}
\end{figure*}

\subsection{Risk Point Identification} \label{III.E}
Unlike the existing CBF method \cite{PCCBF}, which selects the closest point cloud to generate CBFs, we introduce a heuristic approach for identifying high-risk static and dynamic collision points. This method effectively characterizes potential collisions and formulates CBFs to enhance safety by defining points that could potentially drive the robot away from the safe set as risk points.

\emph{Static Risk Point Identification:} Building on the predictive capabilities of NMPC as inspired by \cite{on-demand}, we propose the forward-inverse collision detector, as shown in Fig. \ref{Fig.5}. This detector identifies the first forward and first inverse predicted collisions with the FTD map and selects the corresponding collision point clouds as static risk points. Thanks to the KD-Tree structure, collision detection is easy to implement. Forward detection enables rapid response to front obstacles, while inverse detection prevents the robot from becoming entrapped by single obstacles, making it suitable for multi-obstacle scenarios. We focus on resolving only the most relevant predicted collision, proven to be both effective and computationally efficient. Unlike \cite{on-demand}, our proposed method enforces CBFs as hard collision constraints rather than soft ones. The first predicted collision at time step $k_c$ within the previously considered horizon is defined as
$$
h(\mathbf{x}_{t-1+k_c\vert t-1}) = \Vert \mathbf{x}_{t-1+k_c\vert t-1} - \mathbf{x}_{sta}^*\Vert^2 - \delta_s^2 \le 0,  \eqno{(9)}
$$
where $\delta_s > 0$ represents the minimum static safe distance from the environment. At solving time $t$, the robot has information about the prediction trajectory computed at $t-1$, meaning the collision is predicted to occur at time $t-1+k_c$. Herein, $k_c$ denotes the time step at which the robot predicts a collision with the static local map. The selected static risk point is denoted as $\mathbf{x}_{sta}^*$. To handle complex environments, we add $\mathbf{x}_{sta}^*$ to the historical point set $\mathbf{x}_{his}^*$, which follows an updating principle that removes invalid historical risk points. The final risk description of the static map, represented by the historical point set $\mathbf{x}_{his}^*$, can manage extremely dense environments with an acceptable increase in computational cost.

For each static risk point $\mathbf{x}_{s}^* \in \mathbf{x}_{his}^*$ at each time step $k$, we employ the following static barrier function:
$$
\begin{array}{ccc}
h_{sta}(\mathbf{x}_{t+k\vert t}) = \Vert \mathbf{x}_{t+k\vert t} - \mathbf{x}_{s}^*\Vert^2 - \delta_s^2
\end{array} \eqno{(10)}
$$
for $k=0,1,...,N-1$.

\emph{Dynamic Risk Point Identification:}
As shown in Fig. \ref{Fig.6}, to fully characterize irregular dynamic obstacles, we select dynamic risk points at each prediction time step within the horizon. If the barrier function
$$
\begin{array}{ccc}
    h(\mathbf{x}_{t-1+k\vert t-1}) = \Vert \mathbf{x}_{t-1+k\vert t-1} - \mathbf{x}_{t+k\vert t}^*\Vert^2  - \delta_d^2 \\
     \ge 0, k=0,\ldots, N-1
\end{array} \eqno{(11)}
$$
does not hold, then we select the point $\mathbf{x}_{t+k|t}^* \in P_{dyn,k}$ as the dynamic risk point at each prediction time step $k$. For each dynamic risk point $\mathbf{x}_{t+k|t}^*$ at time step $k$, we employ the following dynamic barrier function:
$$
\begin{array}{ccc}
h_{dyn}(\mathbf{x}_{t+k-1\vert t}) = \Vert \mathbf{x}_{t+k-1\vert t} - \mathbf{x}_{t+k\vert t}^*\Vert^2 - \delta_d^2

\end{array} \eqno{(12)}
$$
where $\delta_d$ is the minimum dynamic safe distance from the environment, with $\delta_d \geq \delta_s> 0$.

The static and dynamic barrier functions $h_{sta}$ and $h_{dyn}$ ensure that the safety set remains forward-invariant, providing a boundary that dynamically responds to both static and dynamic obstacles.

\section{EXPERIMENTS}
In this section, we present the experimental studies conducted in both real-world scenarios and simulation environments to validate the effectiveness of our proposed algorithm. The experiments are carried out under the ROS Noetic operating system. For the dynamic perception module, we configure the local perception range to $5 \times 5$ meters with a resolution of $0.1$ meter, set the dynamic velocity threshold $V_{dyn} = 0.1$ m/s, the dynamic voting threshold $V_{vote} = 0.8$ m/s, and the maximum skip ratio $T_{ratio} = 0.5$. In the local planning module, the horizon length is set to $N = 30$, and the optimization problem is solved using the Casadi solver \cite{casadi}. The local perception module operates at 33 Hz, while the local planning module runs at 10 Hz. Both real-world and simulation experiments utilized the ZED2i RGB-D camera as the perception sensor, providing a wider field of view (FOV) and odometry-based positioning data. All algorithms are executed on a computer equipped with a 2.90 GHz AMD Ryzen 7 4800H CPU and an NVIDIA GTX 1650 GPU.

\begin{figure*}[htbp]
\centering
\includegraphics[width=1\textwidth]{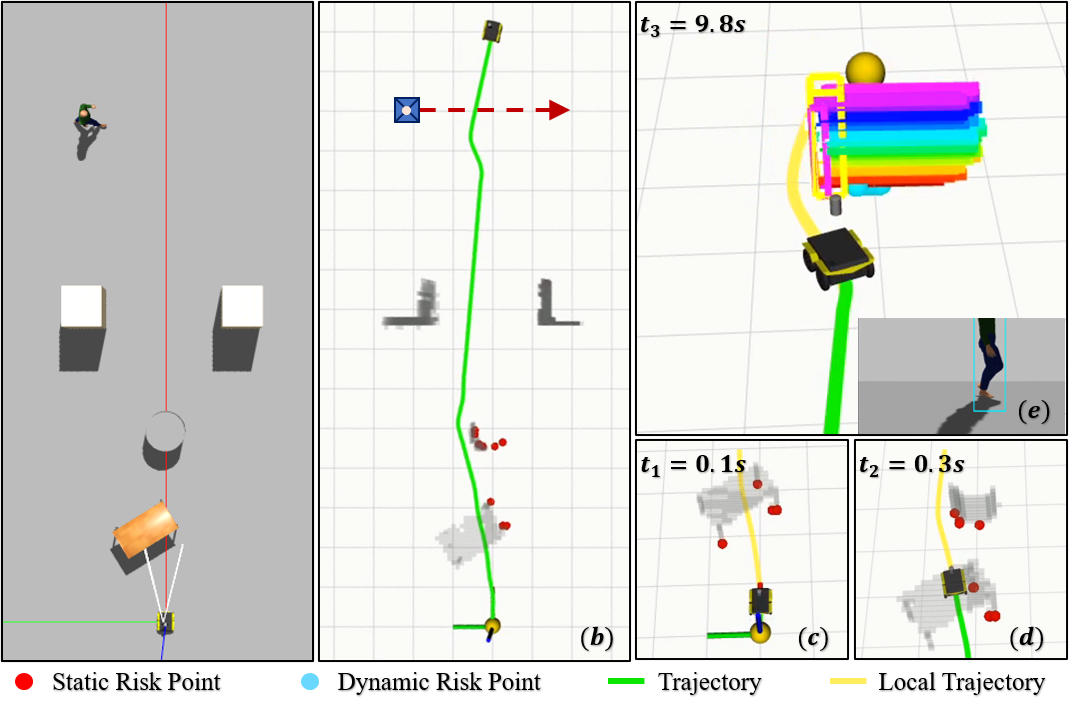}
\caption{Simulation Results. An environment featuring both static and dynamic obstacles is constructed within the Gazebo simulation platform. The autonomous ground robot, Jackal, is employed as the simulated agent. The robot successfully navigates to the target point while avoiding static obstacles (e.g., tables) and dynamic obstacles (e.g., pedestrians). Figs. (a) and (b) illustrate the simulation scenario and the outcomes achieved using our method. Figs (c), (d), and (e) provide detailed insights into the experimental setup and methodological aspects of static and dynamic obstacle avoidance.
}
\label{Fig.8}
\end{figure*}

\subsection{Real-World Scenarios} \label{IV.A}
A real-world experiment is conducted in an unknown indoor environment showed in Fig. \ref{Fig.7}, containing both static and dynamic objects, with limited camera FOV. We select the TurtleBot4 platform for the real-world experiment, setting its speed limits to $0.3$ m/s for linear velocity and $1.2$ rad/s for angular velocity. The distance from the starting point to the target point is set to $8.5$ meters, and the pedestrian's walking speed is approximately $1.2$ m/s. Considering the dimensions of TurtleBot4 and the accuracy of the RGB-D camera, the safe distance $\delta$ is set to $0.271$ meters. The parameter $\gamma$ in the CBF is set to $0.9$.

During the experiment, when TurtleBot4 encounters various static and dynamic obstacles of different shapes and sizes, our algorithm generated a safe trajectory to reach the target point. For instance, in Fig. \ref{Fig.7}, when TurtleBot4 encounters a blackboard-a hollow static obstacle with height restrictions and internal feasible domains, our algorithm successfully identifies the static risk points from the blackboard legs and choose to navigate through the inside rather than bypassing the obstacle, demonstrating less conservative behavior. When encountering pedestrians as dynamic obstacles, our algorithm predicates their future states over the next $N$ steps and successfully identified the dynamic risk points to safely bypass the obstacles. This approach enhances the flexibility and adaptability of trajectory generation.

\subsection{Simulation Scenarios} \label{IV.B}
Fig. \ref{Fig.8} shows the simulation results, showcasing the optimized trajectories in a dynamic environment and a series of snapshots of the Jackal robot. To evaluate the performance of our proposed algorithm, we conducted comparative studies with two baseline methods:
\begin{itemize}
\item DCBF-NMPC \cite{MPC-DCBF}:  An MPC approach that incorporates Kalman filter for obstacle trajectory prediction with uncertainty and a tracking controller that considers safety constraints using a dynamic CBF.
\item Depth-CBF-QP \cite{PCCBF}: A QP-based tracking controller that considers safety constraints using a depth control barrier function.
\end{itemize}
In all comparison studies, we utilize the Jackal robot as our simulation platform, setting its speed limits to $1.2$ m/s for linear velocity and $1.2$ rad/s for angular velocity. The distance from the starting point to the target point is set to $15$ meters, and the pedestrian's walking speed is approximately $1.0$ m/s. We evaluate the performance of each method using five metrics: the minimum distance to obstacles along the path ($\mu_d$), mean relative path length ($\hat{\mu}_l$), mean relative smoothness ($\hat{\mu}_s$), time spent from the start point to the end point ($\mu_t$), and success rate ($\mu$).  The comparison results are summarized in Table \ref{tab.1}.

The Depth-CBF-QP approach fails and collides with obstacles due to its lack of historical and future context, making it challenging to navigate in extremely dense environments. The DCBF-NMPC approach performers poorly with hollow obstacles because it uses ellipses to represent obstacles, leading to unnecessary conservatism in obstacle avoidance. Additionally, the DCBF-NMPC approach fails to identify and cluster irregular obstacles, resulting in failed control barrier function construction and subsequent collisions with objects such as tables. In contrast, our algorithm has successfully avoided all types of static and dynamic obstacles in a smooth and robust manner.

\begin{table}[h]
\caption{Comparison results with two baseline methods$^1$}
\centering
\begin{threeparttable}
  \begin{tabular}{cccccc}\toprule
    \textbf{{Algorithms}} &  {${\mu}_d[m] \downarrow$} & {$\hat{\mu}_l \downarrow$} & {$\hat{\mu}_s \downarrow$} & {${\mu}_t $[s]$\downarrow$} & $\mu\uparrow$ \\\midrule
    \textrm{Ours} & $\mathbf{0.327}$ & $\mathbf{1.000}$ & $\mathbf{1.00}$ & $\mathbf{16.10 }$& $\mathbf{100\%}$ \\
    \textrm{DCBF-NMPC} & 0.609 & 1.086 & 1.98 & 20.75 & $70\%$ \\
    \textrm{Depth-CBF-QP}  & 0 (collided) & - & - & - & $0\%$ \\ \bottomrule
  \end{tabular}
     \begin{tablenotes}
    \footnotesize
    \item[1] $\downarrow$ means the lower that metric, the better the performance, and vice versa. Bold numbers represent the metric-wise best performance.
  \end{tablenotes}
\end{threeparttable}
\label{tab.1}
\end{table}

\section{CONCLUSION}
This paper has presented a safety-critical algorithm for avoiding static and dynamic obstacles using an RGB-D camera. Initially, we constructed an FTD map by integrating the YOLO-Fusion detector with a Kalman filter. Subsequently, risk points were identified from the FTD map, encapsulating both current and predicted future risk information of the environment. Leveraging these risk points, CBFs were formulated and integrated with NMPC to generate a safe, collision-free trajectory. In experimental studies, the proposed algorithm was compared with two baseline methods: DCBF-MPC and Depth-CBF-QP. Experimental results demonstrated that the proposed algorithm offers superior safety and robustness in obstacle avoidance compared to the baseline methods.

\addtolength{\textheight}{-12cm}   



%
%
%

\bibliographystyle{IEEEtran}
\bibliography{IEEEexample}

\end{document}